\pdfoutput=1

\documentclass[11pt]{article}

\usepackage[final]{coling}
\usepackage{times}
\usepackage{latexsym}
\usepackage{booktabs}
\usepackage{multirow}
\usepackage{caption}

\usepackage{tikz}

\usepackage{dblfloatfix}

\usepackage[T1]{fontenc}

\usepackage[utf8]{inputenc}

\usepackage{microtype}

\usepackage{inconsolata}

\usepackage{graphicx}

\usepackage{svg}

\title{Town Hall Debate Prompting: Enhancing Logical Reasoning in LLMs through Multi-Persona Interaction}

\author{Vivaan Sandwar \\
  \texttt{vivaansandwar14@gmail.com} \\\And
  Bhav Jain \\
  \texttt{bhavjain7@gmail.com} \\\And
  Rishan Thangaraj \\
  \texttt{rishanthangaraj@gmail.com} \\
  \bf
  \AND
  Ishaan Garg \\
  \texttt{ishaangarg323@gmail.com} \\\And
  Michael Lam \\
  \texttt{mikelam.us@berkeley.edu} \\\And
  Kevin Zhu \\
  \texttt{zhu502846@berkeley.edu} \\
  }

\begin{document}
\maketitle
\begin{abstract}
Debate is a commonly used form of human communication catered towards problem-solving because of its efficiency. Debate fundamentally allows multiple viewpoints to be brought up in problem-solving, and for complex problems, each viewpoint opens a new path for problem-solving. In this work, we apply this concept to LLM decision-making by proposing town hall-style debate prompting (THDP), a prompting method that splices a language model into multiple personas that will debate one another to reach a conclusion. Our experimental pipeline varies both the number of personas and the personality types of each persona to find the optimum town hall size and personality for benchmark performance as measured by ZebraLogic bench, a reasoning-intensive benchmark characterized by both multiple-choice and fill-in-the-blank questions. Our experimental results demonstrate that a town hall size of 5 personas with LLM-determined personality types performs optimally on ZebraLogic, achieving a 13\% improvement over one-shot CoT baselines in per-cell accuracy in GPT-4o, 9\% puzzle accuracy increase in Claude 3.5 Sonnet, and an improvement in hard puzzle accuracy from 10-15\%.

\end{abstract}

\section{Introduction}

While LLMs are recognized for their impressive abilities in tasks such as text generation, translation, and summarization, they are often relatively less proficient in logical reasoning tasks \cite{intro1,srivastava2023beyond}. This limitation is exposed in problems requiring critical thinking, particularly where a single misstep in the line of reasoning can cascade, resulting in erroneous or hallucinatory responses. Two recent approaches addressing this problem are Chain-of-Thought (CoT) prompting and self-consistency \cite{wei2022chain, selfConsistency}, which work by eliciting the generation of intermediate reasoning steps prior to the final solution. While these prompting techniques have significantly improved the logical reasoning abilities of LLMs, their performance on complex reasoning tasks is still limited \cite{yao2023treethoughtsdeliberateproblem}. One source of this limitation is these techniques' negligence of other lines of reasoning. Prompting techniques like chain-of-thought and self-consistency promote a single, linear path of reasoning, potentially overlooking alternative perspectives that are crucial for tackling more intricate logical tasks \cite{wei2022chain}. Recognizing this gap, we propose a novel prompting technique that simulates the process of a town hall debate. 

\begin{figure}[t]
\includegraphics[scale=0.3]{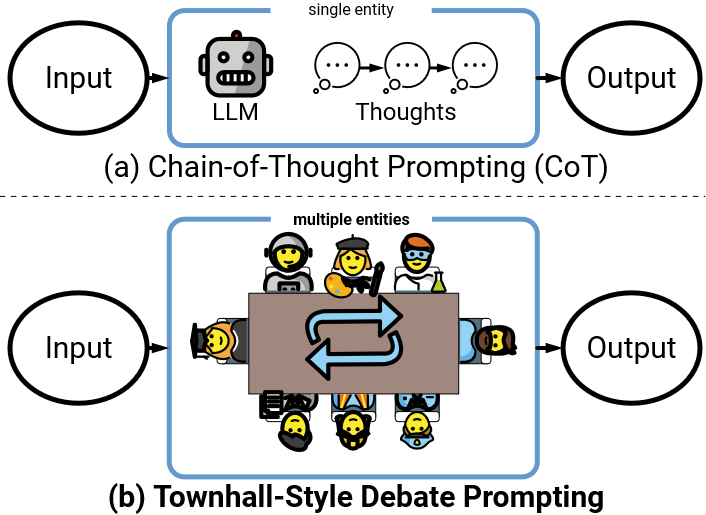}
\caption{Schematic illustration of Townhall-Style Debate Prompting (THDP) and the difference compared to previous prompting methods.}
\end{figure}

This method leverages the concept of a multi-persona debate where we prompt a single LLM to take on multiple “expert” personas across distinct fields and engage in a structured debate over the given problem \cite{wang2024unleashingemergentcognitivesynergy}. Each entity presents its perspective, arguing for or against various solutions, and through this process, they collectively evaluate the merits of each argument. The debate then concludes with a vote, synthesizing the diverse perspectives into a single, logically sound response. A single LLM can thus acquire diverse perspectives and ideas from various domains without additional retrieval and without requiring a commensurate number of instances of separate agents.

Our results showed that the single-model THDP significantly increased model accuracy in Zebragrid Puzzle, Cell, Easy, and Hard evaluation metrics. We achieved our most notable gains with Claude, and overall strongly outperformed CoT prompting, which was used as our control group.

\section{Related Works}

\subsection{Multi-Agent Debate in LLMs}

Recent work relating to multi-persona LLM debate has been implemented using a multi-agent approach. \cite{li2023towardsRW, chen2023survey}
Every agent is assigned a different persona/role, each providing unique reasoning and justification characteristic of its assigned role. These different lines of reasoning are aggregately considered to reach a final solution.
\citet{xu2023improvingRW} demonstrates that this not only increases accuracy on existing benchmarks, but it also enhances the model's ability to solve different kinds of complex reasoning tasks that were not solvable without this approach, due to using a more capable set of LLMs and splitting lines of reasoning. However, THDP specifically calls for divergent thought, and splices a single LLM into multiple personas, allowing for greater in-context memory and more powerful reasoning chains for better answers.

\subsection{LLM Abstract and Logical Reasoning}

Most modern LLMs have not demonstrated the ability to accurately perform tasks that require critical thinking like puzzles and trick questions \cite{zhang2024decouplingProblemSolve}.
\citet{li2024unifiedLogicalReasoning} used LogicAsker – an approach that focuses on evaluating and enhancing reasoning abilities of LLMs under a set of atomic reasoning skills based on different types of logic - to evaluate LLMs performance on logical reasoning problems. They found that there were many logical reasoning failures in a diverse set of LLMs, and the success rates were not consistent at all, ranging from 25\% to 94\%. 

Chain-of-thought has been attempted to mitigate this problem, but improvements have been unimpressive compared to the multi-agent approach \cite{ngyuenLogicalReasoningLimitation}. Inspired by the aforementioned Multi-Agent Debates, we decided to apply that concept to LLMs' ability to solve logical reasoning problems.

\section{Town Hall-Style Debate Prompting}


\subsection{Persona Identification}
THDP first generates a list of $N$ entities with varying personas. 
We let the model dynamically pick these personas to allow meaningful personas for a wide range of tasks. These personas would often be "expert" personas as decided by the model, such as a math, logic, or probability expert. We observe that a state-of-the-art large language model, e.g. GPT-4o can identify accurate and meaningful personas to satisfy this requirement. Through a parameter set at runtime, the user can choose the number of $N$ entities.

\subsection{Multi-Persona Debate}
Utilizing the personas outlined in the previous step, the LLM simulates a Multi-Persona debate where each entity is iterated by describing their logical solution to the problem. Each entity is encouraged to refute problems it may notice in the outputs of other entities, as shown in our prompt, Appendix \ref{sec:prompts}. This debate concludes in 3 rounds of debate, allowing for necessary feedback and rebuttal by each entity.

\subsection{Voting}
After the multi-persona debate has concluded, a voting round begins. Each entity votes on a response that best embodies the final solution and a solution is presented by the majority vote. Each entity is also encouraged to detail why they picked this solution.

\section{Experimental Set-up}

To explore the effectiveness of Townhall-Style Debate Prompting, we employed tasks from reasoning benchmarks that proved challenging even to the most capable LLMs. This evaluation allows for more granular assessment of the prompt in reasoning-intensive domains.

\setlength{\tabcolsep}{3pt}
\begin{table*}[!b]
    \centering
    \caption{THDP vs. CoT on ZebraLogic Grid Prompts}
    \begin{tabular}{lcccccccccc}
        \toprule
        & \multicolumn{2}{c}{Easy} & \multicolumn{2}{c}{Hard} & \multicolumn{2}{c}{Cell} & \multicolumn{2}{c}{Blank} & \multicolumn{2}{c}{Total} \\
        \cmidrule(lr){2-3} \cmidrule(lr){4-5} \cmidrule(lr){6-7} \cmidrule(lr){8-9} \cmidrule(lr){10-11}
        Model & THDP & CoT & THDP & CoT & THDP & CoT & THDP & CoT & THDP & CoT \\
        \midrule
        GPT-4o-mini & 54.5\% & 78.0\% & 2.6\% & 1.9\% & 42.0\% & 41.7\% & 0.0\% & 0.5\% & 14\% & 18.5\% \\
        GPT-4o      & 68.2\% & 75.0 \% & 10\% & 8.2\% & 49.0\% & 36.0\% & 0.8\% & 22.3\% & 24\% & 22.3\% \\
        Claude-3.5-Sonnet & 86.4\% & 84.1\% & 15.4\% & 10.9\% & 54.8\% & 53.1\% & 0.3\% & 0.5\% & 37.0\% & 28.9\% \\
        \bottomrule
    \end{tabular}
    \caption*{THDP compared to Chain of Thought for the ZebraLogic Grid Benchmark, under 5 personas. Results show the various improvements gained by using THDP. Results also show how improvements improve as models scale.  }
    \label{table:grid}
\end{table*}

\subsection{Zebra Logic Bench}
\subsubsection{Task Description}
We utilize the \textbf{Zebra Logic Bench} created by ZebraLogic \cite{zebralogicbench2024,dziri2024faith}. In this benchmark, each question involves filling out an NxM grid based on a list of clues. N corresponds to the number of "Zebra Houses", and M is the "Zebra Feature" number. The difficulty of the problem grows as the product of N and M. This benchmark is split into two different subsets: multiple-choice questions (MCQ) and ZebraGrid. In MCQ, the LLM must pick the letter corresponding to the correct grid assignment. ZebraGrid increases the difficulty by requiring the model to output the filled grid itself. As noted by ZebraLogic, the ZebraGrid subset has a substantially lower percent solve rate by even state-of-the-art large language models such as Claude 3.5 Sonnet. Based on budget considerations, we selected the first 200 rows of each subset for evaluation.
\subsubsection{Evaluation Metrics}
For the MCQ subset, we calculated how many were answered correctly and incorrectly, including the outputs that couldn't be understood, and labeled as blanks. For the ZebraGrid subset, we also added a metric tracking a percentage of how many harder problems and easy problems were answered correctly as well as overall cell accuracy, measured as  $\displaystyle\frac{\mbox{\# of correct cells}}{\mbox{\# of total cells}}$.

\subsection{Model Choice}
The models on which we tested our approach include GPT-4o, GPT-4o Mini, and Claude 3.5 Sonnet. Claude 3.5 Sonnet was selected as it was, at the time, one of the current state-of-the-art large language models \cite{chatbotArena}, while GPT-4o Mini was chosen to determine the effects of our prompt on a smaller model.

\subsubsection{Baselines}
We compare our approach to prompting with \textbf{1-Shot Chain-Of-Thought (COT)} prompting. Prompt descriptions for the methods can be found in Appendix~\ref{sec:prompts}.

\section{Results}


\begin{figure*}[t]
\includegraphics[scale=0.4]{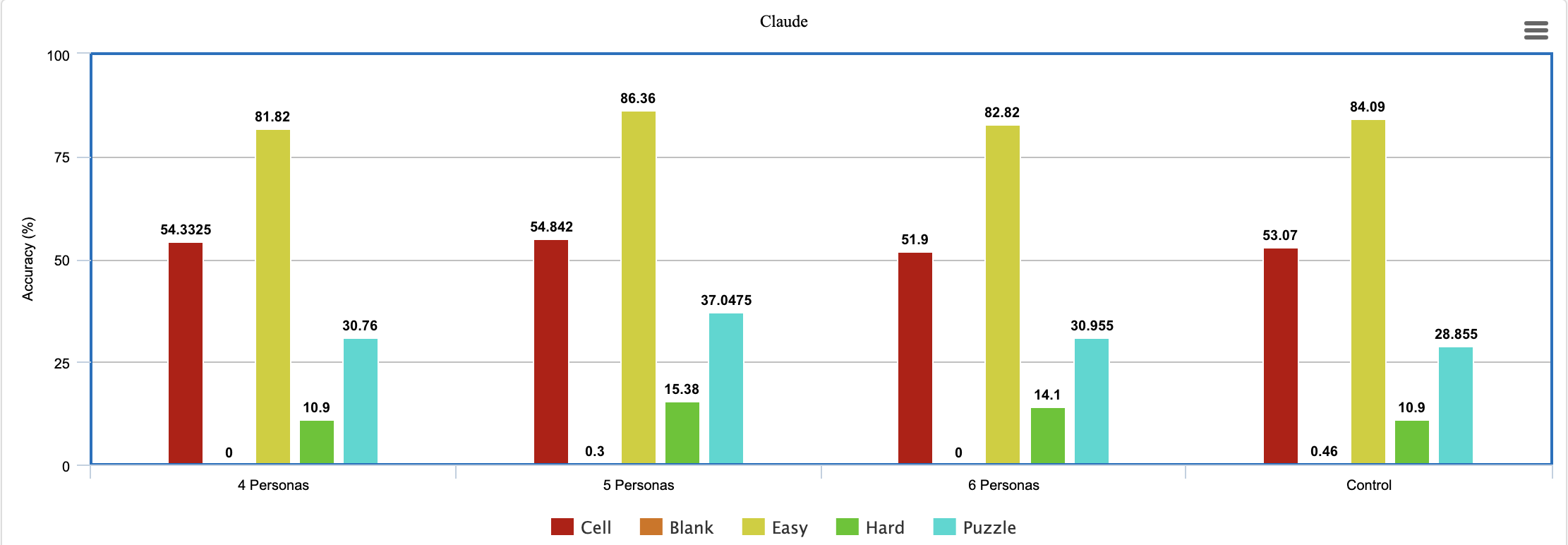}
\caption{By using various persona counts, we can see how model output varies and performs. We note that a persona count of 5 performs generally the best. }
\label{fig:claude}
\end{figure*}

\begin{figure}[t]
\includegraphics[scale=0.18]{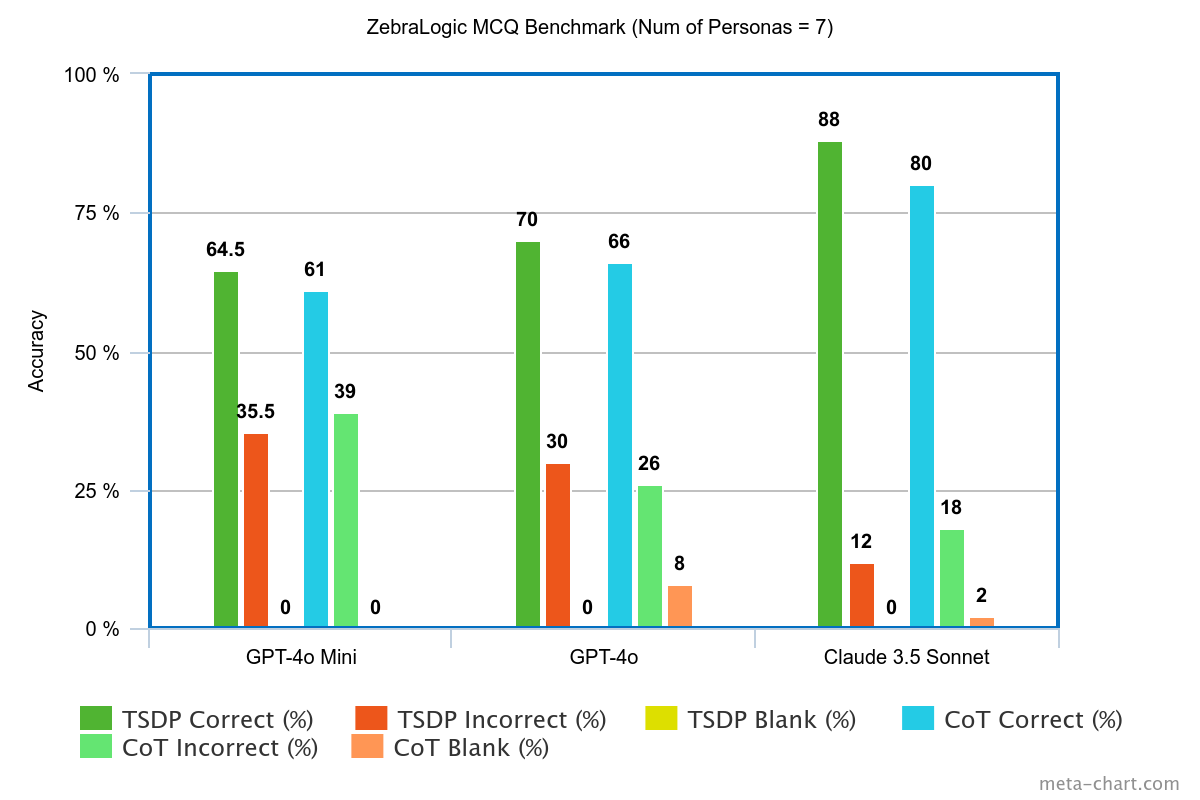}
\caption{THDP demonstrates better results across the board on MCQ-style problems. Higher correct is better, lower incorrect and blank is better.}
\label{fig:mcq}
\end{figure}

Table \ref{table:grid} and Figure \ref{fig:mcq} show results from the ZebraLogic benchmark, comparing our prompting technique to CoT prompting across ZebraGrid and Zebra MCQ evaluations. Our technique showed significant gains over CoT prompting, particularly in Cell accuracy and Hard puzzle accuracy for GPT-4o, GPT-4o Mini, and Claude 3.5 Sonnet.

We conducted over 30 individual evaluations with persona counts ranging from 2 to 15, with final testing using 4, 5, and 6 personas. The 5-persona groups consistently performed best.

Referring to Figure \ref{fig:claude}, for Claude 3.5 Sonnet:
\begin{itemize}
\item 4 persona group: Easy puzzle accuracy slightly worse than control, but cell and puzzle accuracy 1.3-2\% better.
\item 5 persona group: Hard puzzle accuracy improved 1.5x (10\% to 15\%), overall puzzle accuracy increased by 8\% (29\% to 37\%).
\end{itemize}

For GPT-4o (Figure \ref{fig:mcq}):
\begin{itemize}
\item 4 persona group: Cell accuracy improved by almost 13\%, blank occurrences reduced from 22.5\% to 0\%. Easy puzzle accuracy 5\% worse, but hard puzzle accuracy 2\% better, and overall puzzle accuracy 1.25\% higher.
\end{itemize}

GPT-4o-mini results:
\begin{itemize}
\item 4 personas: Cell accuracy 1.67\% lower, Easy accuracy 23\% lower, Hard accuracy increased 1.6x, overall puzzle accuracy 4\% lower.
\item 5 personas: Cell accuracy 0.33\% higher, Easy accuracy 23\% lower, Hard accuracy 1.3x higher.
\item 6 personas: Similar to 5 personas, with slightly lower cell accuracy.
\end{itemize}

The weaker performance of GPT-4o Mini is likely due to internal debates diverging without error corrections, as discussed in the discussion section.

\subsection{MCQ Results}
For MCQ evaluations, 7 personas performed best:
\begin{itemize}
\item GPT-4o-mini: Accuracy improved by 3.5\%
\item GPT-4o: Accuracy improved by 4\%, blank rate reduced from 8\% to 0\%
\item Claude 3.5: Accuracy increased by 8\%, blank rate reduced from 2\% to 0\%
\end{itemize}

\section{Discussion}
THDP effectively improves the output of reasoning-intensive tasks as opposed to other prompting techniques such as Chain of Thought reasoning. As demonstrated by the results in section 3, Townhall-Style Debate Prompting brings significant improvements to reasoning-intensive tasks such as the ZebraLogic benchmark. 

 Unlike chain-of-thought reasoning, this technique effectively amplifies its effects by bringing together the reasoning of multiple entities, which results in better reasoning abilities. By allowing the LLM to debate itself diverse opinions, counterarguments, and complicated, multi-step solutions also appeared. By the final round, these counterarguments and potential solutions would help lead to the best solution possible, created by the rounds of debate amongst the various entities. By analyzing multiple viewpoints, a more nuanced understanding of the situation exists which leads to a better answer as various entities come together.

\subsection{Conclusions}
The performance of stronger/larger models like Claude 3.5 Sonnet and GPT-4o improves much more as a result of THDP than smaller models such as GPT-4o Mini. A potential cause of this may be that these smaller models might not have the capability to fully carry out such a task. The Townhall Debate Prompt is more generally an amplifier of LLM capabilities, for a more powerful base LLM, it essentially allows itself to effectively utilize divergent thought to find answers to more difficult questions, but for weaker LLMs, such as GPT 4o Mini, the internal debate can and often did lead to the LLMs going off on tangents, and more often getting stuck within a repetitive loop of incorrect thoughts with minimal error correction occurring.

\subsection{Limitations}
Although Townhall-Style Debate Prompting exhibits significant improvements in reasoning capability, it has some limitations. For example, even with proper personas being chosen and increased reasoning capability, the answer isn't guaranteed to be correct. It also remains unclear whether a certain type of persona exhibits greater results than other types. For example, it is unknown whether a domain-specific persona exhibits better reasoning than one that plays the role of a certain persona, such as the "Devil's Advocate." Future experimentation is needed to quantify the impact of having a certain persona or not.

Furthermore, due to the nature of the prompt, more tokens are inevitably used and generated in model response. This results in increased computation and model costs as well as longer response times. This token-intensive approach also causes issues with models that might have a token limit that is too low, causing the response to effectively cut off.

\bibliography{custom}

\begin{thebibliography}{15}
\providecommand{\natexlab}[1]{#1}

\bibitem[{bench authors(2023)}]{srivastava2023beyond}
BIG bench authors. 2023.
\newblock \href {https://openreview.net/forum?id=uyTL5Bvosj} {Beyond the imitation game: Quantifying and extrapolating the capabilities of language models}.
\newblock \emph{Transactions on Machine Learning Research}.

\bibitem[{Bill Yuchen~Lin(2024)}]{zebralogicbench2024}
Yejin~Choi Bill Yuchen~Lin, Ronan Le~Bras. 2024.
\newblock \href {https://hf.co/spaces/allenai/ZebraLogicBench-Leaderboard} {Zebralogic: Benchmarking the logical reasoning ability of language models}.

\bibitem[{Dziri et~al.(2024)Dziri, Lu, Sclar, Li, Jian, Lin, West, Bhagavatula, Bras, Hwang, Sanyal, Welleck, Ren, Ettinger, Harchaoui, and Choi}]{dziri2024faith}
Nouha Dziri, Ximing Lu, Melanie Sclar, Xiang~Lorraine Li, Liwei Jian, Bill~Yuchen Lin, Peter West, Chandra Bhagavatula, Ronan~Le Bras, Jena~D. Hwang, Soumya Sanyal, Sean Welleck, Xiang Ren, Allyson Ettinger, Za{"i}d Harchaoui, and Yejin Choi. 2024.
\newblock \href {https://arxiv.org/abs/2305.18654} {Faith and fate: Limits of transformers on compositionality}.
\newblock \emph{Advances in Neural Information Processing Systems}, 36.

\bibitem[{Ha-Thanh~Nguyen(2023)}]{ngyuenLogicalReasoningLimitation}
Ken~Satoh Ha-Thanh~Nguyen, Wachara~Fungwacharakorn. 2023.
\newblock \href {https://arxiv.org/abs/2311.13095} {Enhancing logical reasoning in large language models to facilitate legal applications}.

\bibitem[{Jack W.~Rae(2021)}]{intro1}
Trevor~Cai Jack W.~Rae, Sebastian~Borgeaud. 2021.
\newblock \href {https://arxiv.org/abs/2112.11446} {Scaling language models: Methods, analysis \& insights from training gopher}.

\bibitem[{Jason~Wei(2022)}]{wei2022chain}
Dale~Schuurmans Jason~Wei, Xuezhi~Wang. 2022.
\newblock \href {https://arxiv.org/abs/2201.11903} {Chain-of-thought prompting elicits reasoning in large language models}.
\newblock \emph{arXiv preprint arXiv:2201.11903}.

\bibitem[{Rasal(2024)}]{zhang2024decouplingProblemSolve}
Sumedh Rasal. 2024.
\newblock \href {https://arxiv.org/abs/2401.01312} {Llm harmony: Multi-agent communication for problem solving}.
\newblock \emph{arXiv preprint arXiv:2401.01312}.

\bibitem[{Tian~Liang(2023)}]{li2023towardsRW}
Wenxiang~Jiao Tian~Liang, Zhiwei~He. 2023.
\newblock \href {https://arxiv.org/abs/2305.19118} {Encouraging divergent thinking in large language models through multi-agent debate}.
\newblock \emph{arXiv preprint arXiv:2305.19118}.

\bibitem[{Wang et~al.(2024)Wang, Mao, and Wu}]{wang2024unleashingemergentcognitivesynergy}
Zhenhailong Wang, Shaoguang Mao, and Wenshan Wu. 2024.
\newblock \href {https://arxiv.org/abs/2307.05300} {Unleashing the emergent cognitive synergy in large language models: A task-solving agent through multi-persona self-collaboration}.
\newblock \emph{Preprint}, arXiv:2307.05300.

\bibitem[{Wei-Lin~Chiang(2024)}]{chatbotArena}
Ying~Sheng Wei-Lin~Chiang, Lianmin~Zheng. 2024.
\newblock \href {https://arxiv.org/abs/2403.04132} {Chatbot arena: An open platform for evaluating llms by human preference}.

\bibitem[{Xuezhi~Wang(2022)}]{selfConsistency}
Dale~Schuurmans Xuezhi~Wang, Jason~Wei. 2022.
\newblock \href {https://arxiv.org/abs/2203.11171} {Self consistency improves chain of thought reasoning in language models}.

\bibitem[{Yao et~al.(2023)Yao, Yu, Zhao, Shafran, Griffiths, Cao, and Narasimhan}]{yao2023treethoughtsdeliberateproblem}
Shunyu Yao, Dian Yu, Jeffrey Zhao, Izhak Shafran, Thomas~L. Griffiths, Yuan Cao, and Karthik Narasimhan. 2023.
\newblock \href {https://arxiv.org/abs/2305.10601} {Tree of thoughts: Deliberate problem solving with large language models}.

\bibitem[{Yuxuan~Wan(2024)}]{li2024unifiedLogicalReasoning}
Yiliu~Yang Yuxuan~Wan, Wenxuan~Wang. 2024.
\newblock \href {https://arxiv.org/abs/2401.00757} {Triggering logical reasoning failures in large language models}.
\newblock \emph{arXiv preprint arXiv:2401.00757}.

\bibitem[{Zhenhailong~Wang(2023)}]{xu2023improvingRW}
Wenshan~Wu Zhenhailong~Wang, Shaoguang~Mao. 2023.
\newblock \href {https://arxiv.org/abs/2307.05300} {Unleashing the emergent cognitive synergy in large language models: A task-solving agent through multi-persona self-collaboration}.
\newblock \emph{arXiv preprint arXiv:2307.05300}.

\bibitem[{Zhiheng~Xi(2023)}]{chen2023survey}
Xin~Guo Zhiheng~Xi, Wenxiang~Chen. 2023.
\newblock \href {https://arxiv.org/abs/2309.07864} {The rise and potential of large language model based agents: A survey}.
\newblock \emph{arXiv preprint arXiv:2309.07864}.

\end{thebibliography}

\appendix

\section{Prompts}
\label{sec:prompts}

To instantiate the LLM to perform a multi-persona debate that follows the expected structure, we carefully designed the structure of the thdp prompt as follows.
\subsection{System Principle}
The first part of the prompt contains a high-level-instruction:
\texttt{When faced with a task, begin by identifying the {n} entities who will contribute to solving the task. Then, initiate a 3 round townhall-style debate process, and vote upon the final response after the 3 rounds. Each entity should give critical comments and rebuttals whenever necessary, along with their responses to the task.} \textit{n} represents the number of entities to pick, which is chosen at run-time.

\subsection{Demonstration Examples}
Then, we include 1-2 manually written examples to showcase the expected debate behavior, depending on which dataset is run on, MCQ or the Grid Puzzle, to reflect the default COT prompt provided for these puzzles, which already include a COT example. This example assists the LLM in determining how to structure the debate as well as output and provides a reference point in providing solutions. 

\subsection{Task Prefix}
The final part of the prompt is to remind the LLM to identify a panel of entities and collaboratively solve the task with the correct number of entities, followed by a task-specific formatting guide. 

\section{Inference Configurations}
GPT 4o and 4o Mini were used through the default chat completions API provided by OpenAI. Claude 3.5 Sonnet was accessed through Anthropic's API as well. For both OpenAI models, we utilized a greedy sampling technique. For Claude 3.5 Sonnet, we utilized a temperature sampling of 0.5 which generally yielded better results on the default prompt and THDP based on experimentation.  

\section{ZebraLogic Benchmark}
\begin{figure}[h]
\includegraphics[scale=0.3]{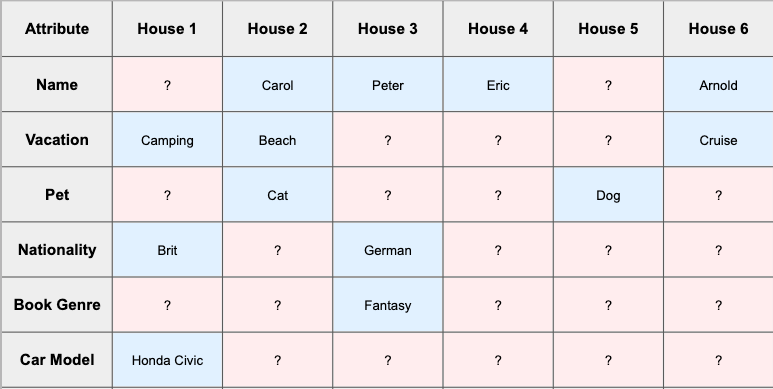}
\caption{Visual depiction of what a grid benchmark task might look like in this case, specifically the ZebraLogic Benchmark}
\end{figure}

\begin{figure*}[t]
\centering
\includegraphics[scale=0.4]{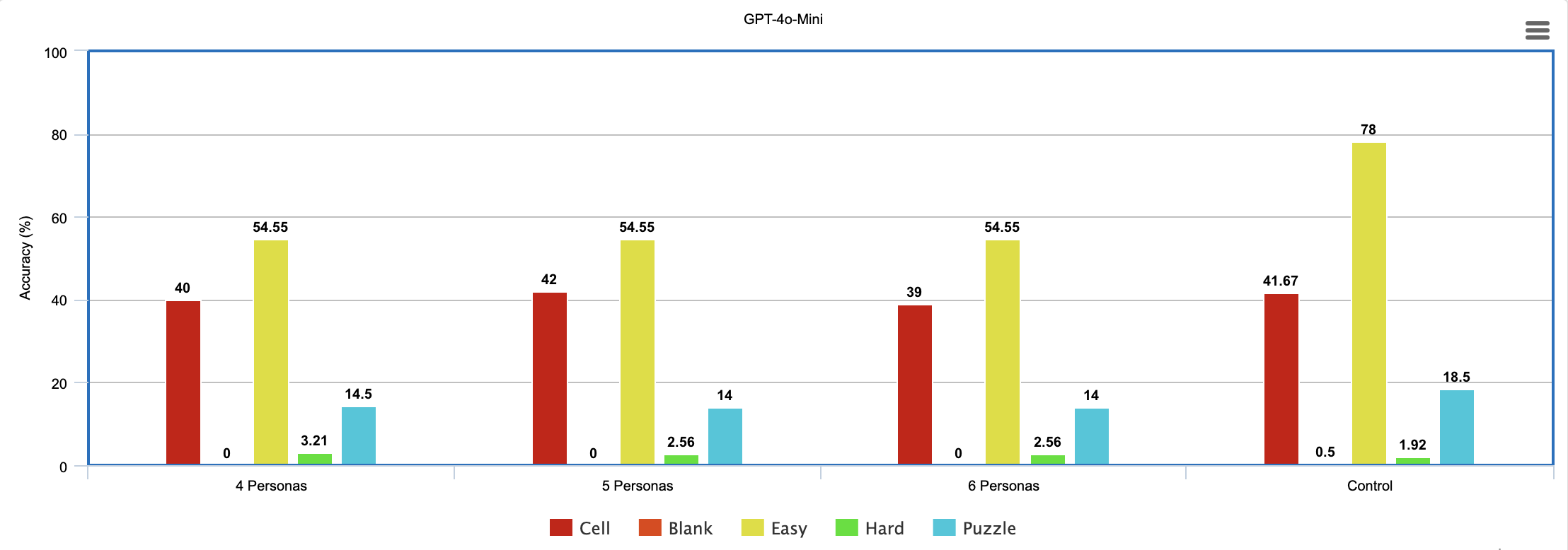}
\caption{THDP shows weaker results on smaller models such as GPT-4o-Mini. Higher Cell, Easy, Hard, and Puzzle accuracies are better. A lower blank is better.}
\end{figure*}
\begin{figure*}[t]
\includegraphics[scale=0.4]{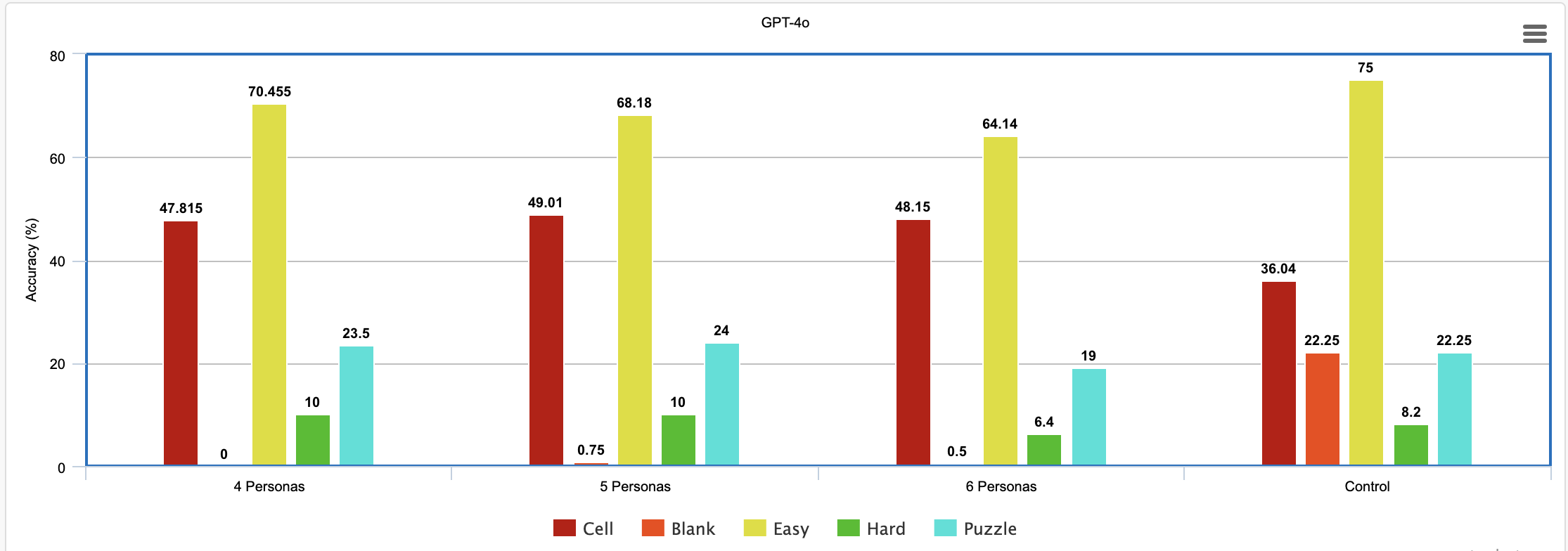}
\caption{THDP demonstrates improved performance on larger models such as GPT-4o, improving metrics such as Cell Accuracy, Hard Problem Accuracy, and Blank Accuracy. Higher Cell, Easy, Hard, and Puzzle accuracies are better. A lower blank is better.}
\end{figure*}

\clearpage

\begin{tikzpicture}[remember picture, overlay]
  \draw[thick, rounded corners=15pt]
    ([shift={(1in,-1in)}]current page.north west) rectangle ([shift={(-1in,1in)}]current page.south east);
    \node[anchor=north west, text width=\paperwidth-2in, align=center] at ([shift={(1in,-1in)}]current page.north west)
    {\Large MCQ Prompt};
    \node[anchor=south west, text width=\paperwidth-3in, align=left] at ([shift={(1.5in,1.5in)}]current page.south west)
    {\footnotesize
    When faced with a task, begin by identifying the \{n\} entities who will contribute to solving the task. Then, initiate a 3 round townhall-style debate process, and vote upon the final response after the 3 rounds. Each entity should give critical comments and rebuttals whenever necessary, along with their responses to the task.

Here are some examples. Note that these examples use any amount of entities, while you will have to use \{n\}.:
—
Puzzle: In a small neighborhood, there are four houses in a row, numbered 1 to 4 from left to right. Each house is owned by a person of a different nationality (American, British, Canadian, Dutch), has a different colored roof (Red, Blue, Green, Yellow), and has a different pet (Dog, Cat, Fish, Bird). Use the following clues to determine the details of each house:
The American lives in the house with the Red roof.
The Dog owner lives immediately to the right of the house with the Blue roof.
The Canadian lives in the rightmost house.
The Bird owner lives next to the house with the Yellow roof.
The British person owns a Cat.
The house with the Green roof is somewhere to the left of the house with the Yellow roof.
Question: Who owns the Fish?
Choices: A) The American B) The British C) The Canadian D) The Dutch
Reasoning:
Round 1:
Logical Laura (Expert in deductive reasoning): Let's start by organizing the given information. We have 4 houses, 4 nationalities, 4 roof colors, and 4 pets. From clue 3, we know the Canadian is in house 4. Clue 1 tells us the American has a Red roof, but they can't be in house 4, so they must be in house 1, 2, or 3. Let's start filling in what we know and work from there.
Systematic Sam (Specialist in problem-solving methodologies): Good start, Laura. I agree with your initial deductions. Let's create a table to visualize what we know so far:
House | Nationality | Roof Color | Pet 1 | ? | ? | ? 2 | ? | ? | ? 3 | ? | ? | ? 4 | Canadian | ? | ?
We should also note that the British person owns a Cat (clue 5), and they can't be in house 4.
Intuitive Ivy (Pattern recognition expert): Excellent groundwork. I notice that clue 6 about the Green and Yellow roofs, combined with the Canadian being in house 4, means the Yellow roof must be on house 3 or 4, and the Green roof must be on house 1 or 2. This, combined with the American having a Red roof, leaves only one possibility for the Blue roof - it must be on the house next to the Dog owner (clue 2).
Round 2:
Logical Laura: Great observations. Let's continue our deductions. Since the Dog owner is to the right of the Blue roof, and the Canadian (in house 4) isn't mentioned as owning any specific pet, the Dog must be in house 3, and the Blue roof must be on house 2. This means the American with the Red roof must be in house 1.
Systematic Sam: Excellent deduction, Laura. Let's update our table:
House | Nationality | Roof Color | Pet 1 | American | Red | ? 2 | ? | Blue | ? 3 | ? | ? | Dog 4 | Canadian | ? | ?
We know the British person owns a Cat and can't be in house 4, so they must be in house 2.
Intuitive Ivy: Good progress! Remember, the Bird owner lives next to the Yellow roof (clue 4). Since we know the Yellow roof is on house 3 or 4, and house 3 has the Dog, the Bird must be in house 4 with the Canadian. This leaves the Fish for either house 1 or 2.
Round 3:
Logical Laura: Great catch, Ivy. Now we can almost complete our table. The Dutch person must be in house 3 with the Dog. The only pet left unassigned is the Fish, and the only house without a pet is house 1. Therefore, the American in house 1 must own the Fish.
Systematic Sam: I agree with Laura's conclusion. Let's see our final table:
House | Nationality | Roof Color | Pet 1 | American | Red | Fish 2 | British | Blue | Cat 3 | Dutch | Green | Dog 4 | Canadian | Yellow | Bird
This arrangement satisfies all the given clues.
Intuitive Ivy: I concur with this final arrangement. It fits all the clues and our logical deductions. The American owns the Fish, which corresponds to choice A in our options.
Voting:
Logical Laura: I vote for option A. Our step-by-step deduction clearly shows the American owns the Fish. Systematic Sam: I also vote for option A. Our systematic approach and final table support this conclusion. Intuitive Ivy: I agree with option A. The pattern we uncovered through our analysis points to the American as the Fish owner.
Summary: All three experts unanimously agree that the American owns the Fish. This conclusion was reached through a careful analysis of the given clues, systematic elimination, and logical deduction. The process involved creating a visual representation of the information and methodically filling in the details based on the given clues and their implications.
Answer: A

—

\#\# Puzzle:

\{puzzle\}

\#\# Question:

\{question\}

\#\# Choices:

\{choices\}

\#\# Instruction

Please answer this question by first debating as shown above.
It is very important to present your reasoning and solution in the following json format.
It is also important to show your choice in the `answer` field with only the choice letter, e.g.,`"answer": "C"`.
It is imperative you follow this json format below, with nothing outside of it.

\{json\_template\}
    
};
\end{tikzpicture}

\clearpage

\begin{tikzpicture}[remember picture, overlay]
  \draw[thick, rounded corners=15pt]
    ([shift={(1in,-1in)}]current page.north west) rectangle ([shift={(-1in,1in)}]current page.south east);
    \node[anchor=north west, text width=\paperwidth-2in, align=center] at ([shift={(1in,-1in)}]current page.north west)
    {\Large Grid Prompt};
    \node[anchor=south west, text width=\paperwidth-3in, align=left] at ([shift={(1.5in,1.5in)}]current page.south west)
    {\footnotesize
\#\#\# Instructions
When faced with a task, begin by identifying the \{n\} entities who will contribute to solving the task. Then, initiate a 3 round townhall-style debate process, and vote upon the final response after the 3 rounds. Each entity should give critical comments and rebuttals whenever necessary, along with their responses to the task.

Debate Process:

Have each AI entity present their initial thoughts on the topic using chain-of-thought reasoning. They should explicitly walk through their logical steps.
Allow for 3 rounds of rebuttals and counter-arguments, ensuring each entity maintains consistency with their established viewpoint and expertise.
Encourage entities to refine or adjust their stances based on new information presented during the debate, while still maintaining overall consistency with their core principles.

Here are some examples. Note that these examples use any amount of entities, while you will have to use \{n\}.:
—
\#\#\# Example 1 -- Easy problems

Problem
There are 2 houses, numbered 1 to 2 from left to right, as seen from across the street. Each house is occupied by a different person. Each house has a unique attribute for each of the following characteristics:
Each person has a unique name: Eric, Arnold
Each person has a unique type of pet: dog, cat
Clues:
Eric is somewhere to the left of Arnold.
The person who owns a dog is not in the first house.
Debate Style Reasoning:
Round 1: Initial Observations
Logical Deduction Expert:
Let's start with the clues we have. The first clue is that Eric is somewhere to the left of Arnold. Since we only have two houses, this means Eric must be in House 1 and Arnold must be in House 2.
Data Organizer:
Based on the deduction from the Logical Deduction Expert, we can fill out part of the information:
House 1: Eric
House 2: Arnold
Next, we need to determine the pets.
Pattern Recognition Specialist:
From Clue 2, we know that the person who owns a dog is not in the first house. Since Eric is in House 1, he cannot have the dog. Therefore, Eric must have the cat. Consequently, Arnold, being in House 2, must have the dog.
Constraint Satisfaction Solver:
Let’s verify:
Eric is to the left of Arnold, which fits the clue.
The person with the dog is not in the first house, so Arnold in House 2 having the dog fits the clue.
Everything aligns with the clues and constraints provided.
Hypothesis Tester:
We should test the reverse scenario for thoroughness. Suppose Eric were in House 2, then Arnold would be in House 1. This would place Arnold to the left of Eric, which violates the first clue. Also, if Eric were in House 2 and had the dog, it would violate the second clue. Thus, the initial arrangement where Eric is in House 1 and Arnold in House 2 with the cat and dog respectively is correct.
Sequence Analyst:
From the clues and our filled-in information:
Eric (House 1) has to be to the left of Arnold (House 2).
The dog must be in House 2, which matches our arrangement.
All sequences and constraints are satisfied.
Final Answer:
House 1: Eric, Cat
House 2: Arnold, Dog

\#\#\# Example 2 -- hard problems

Puzzle: There are 6 houses, numbered 1 to 6 from left to right, as seen from across the street. Each house is occupied by a different person. Each house has a unique attribute for each of the following characteristics:
- Each person has a unique name: Alice, Carol, Eric, Peter, Bob, Arnold
- People have unique favorite music genres: classical, hip hop, jazz, pop, rock, country
- The mothers' names in different houses are unique: Sarah, Penny, Aniya, Janelle, Kailyn, Holly
- Each mother is accompanied by their child: Alice, Fred, Timothy, Bella, Samantha, Meredith
- People have unique heights: very short, tall, short, very tall, super tall, average
- The people keep unique animals: bird, dog, horse, rabbit, cat, fish

Clues:
1. The person who loves pop music is the cat lover.
2. The rabbit owner is directly left of The person whose mother's name is Aniya.
3. The person whose mother's name is Holly is directly left of Carol.
4. The person whose mother's name is Holly is the person's child is named Alice.
5. The person whose mother's name is Holly is the person who loves classical music.
6. The person who loves jazz music is The person whose mother's name is Sarah.
7. The person's child is named Meredith is somewhere to the right of The person whose mother's name is Aniya.
8. The person who is super tall is The person whose mother's name is Holly.
9. The person who is the mother of Timothy is Bob.
10. The person who is very short is somewhere to the left of The person whose mother's name is Aniya.
11. Eric is the fish enthusiast.
12. The person's child is named Samantha is somewhere to the right of the person who is very tall.
13. The person who loves rock music is The person whose mother's name is Janelle.
14. There is one house between the person who keeps horses and the person's child is named Meredith.
15. The person's child is named Bella is somewhere to the right of Peter.
16. The fish enthusiast is somewhere to the left of the bird keeper.
17. The fish enthusiast is somewhere to the right of the person's child is named Alice.
18. There is one house between the person's child is named Bella and the person who loves rock music.
19. The person who is short is the cat lover.
20. Alice is directly left of the person who loves classical music.
21. The person's child is named Bella is The person whose mother's name is Aniya.
22. There are two houses between The person whose mother's name is Penny and the person who is short.
23. The person who loves hip-hop music is in the first house.
24. Carol is the person who is tall.

—-
};
\end{tikzpicture}

\clearpage

\begin{tikzpicture}[remember picture, overlay]
  \draw[thick, rounded corners=15pt]
    ([shift={(1in,-1in)}]current page.north west) rectangle ([shift={(-1in,1in)}]current page.south east);
    \node[anchor=north west, text width=\paperwidth-2in, align=center] at ([shift={(1in,-1in)}]current page.north west)
    {\Large Grid Prompt};
    \node[anchor=south west, text width=\paperwidth-3in, align=left] at ([shift={(1.5in,1.5in)}]current page.south west)
    {\footnotesize
 
Question: What is the complete arrangement of the houses and their attributes?

Personas:

1. Logical Deduction Expert: Skilled in drawing conclusions from given information and identifying logical relationships.
2. Data Organizer: Proficient in structuring and visualizing complex information in tables or matrices.
3. Pattern Recognition Specialist: Adept at identifying recurring patterns and relationships within data sets.
4. Constraint Satisfaction Solver: Expert in resolving problems with multiple interrelated constraints and requirements.
5. Sequence Analyst: Specialized in determining optimal arrangements and sequences based on given rules.
6. Hypothesis Tester: Skilled at formulating and systematically testing hypotheses based on available information.

\{
“
Reasoning:

Round 1:

Logical Deduction Expert:
Let's start by analyzing the direct information provided in the clues:
1. The person who loves hip-hop music is in the first house (Clue 23).
2. Carol is the person who is tall (Clue 24).
3. Eric is the fish enthusiast (Clue 11).
4. The person who loves pop music is the cat lover (Clue 1).
5. Bob is the mother of Timothy (Clue 9).
6. The person who loves jazz music is the person whose mother's name is Sarah (Clue 6).
7. The person who loves rock music is the person whose mother's name is Janelle (Clue 13).
8. The person whose mother's name is Holly is the person who loves classical music (Clue 5).
9. The person whose mother's name is Holly is super tall (Clue 8).
10. The person whose mother's name is Holly is the person whose child is named Alice (Clue 4).
11. The person who is short is the cat lover (Clue 19).
12. Alice is directly left of the person who loves classical music (Clue 20).
13. The person's child named Bella is the person whose mother's name is Aniya (Clue 21).

Data Organizer:
Based on the initial deductions, let's create a partial table to visualize what we know:

House | Name | Music Genre | Mother | Children | Height | Animal
1     | ?    | hip hop     | ?      | ?        | ?      | ?
2     | ?    | ?           | ?      | ?        | ?      | ?
3     | ?    | ?           | ?      | ?        | ?      | ?
4     | ?    | ?           | ?      | ?        | ?      | ?
5     | Eric | ?           | ?      | ?        | ?      | fish
6     | ?    | ?           | ?      | ?        | ?      | ?

Pattern Recognition Specialist:
I notice several patterns in the clues that can help us determine relative positions:
1. The rabbit owner is directly left of the person whose mother's name is Aniya (Clue 2).
2. The person whose mother's name is Holly is directly left of Carol (Clue 3).
3. The person's child named Meredith is somewhere to the right of the person whose mother's name is Aniya (Clue 7).
4. The person who is very short is somewhere to the left of the person whose mother's name is Aniya (Clue 10).
5. The person's child named Samantha is somewhere to the right of the person who is very tall (Clue 12).
6. The person's child named Bella is somewhere to the right of Peter (Clue 15).
7. The fish enthusiast is somewhere to the left of the bird keeper (Clue 16).
8. The fish enthusiast is somewhere to the right of the person's child named Alice (Clue 17).
9. There is one house between the person's child named Bella and the person who loves rock music (Clue 18).
10. There are two houses between the person whose mother's name is Penny and the person who is short (Clue 22).

Constraint Satisfaction Solver:
Let's apply some constraints to narrow down our options:
1. Since Eric (fish enthusiast) is in house 5, the bird keeper must be in house 6.
2. The person whose child is Alice must be in houses 1-4, as Eric is to the right of them.
3. The rabbit owner must be in houses 1-5, as they are directly left of someone.
4. Carol (tall) must be in houses 2-6, as someone is directly left of her.
5. The person who is very short must be in houses 1-4, as they are to the left of the person whose mother's name is Aniya.
6. The person who loves pop music (cat lover) is short, so they can't be the same as the very short person.
7. There must be at least two houses between the very short person and the short person (cat lover).

Sequence Analyst:
Based on the patterns and constraints, let's try to establish some sequences:
1. Alice (hip hop) - ? - ? - Bob (pop, short, cat) - Eric (fish) - Arnold (jazz, bird)
2. The sequence: very short - ? - Aniya's child (Bella) - ? - rock music lover
3. The sequence: Penny's child - ? - ? - short person (cat lover)
4. The sequence: person with Alice as child - Carol (tall) - ? - ? - Eric - Arnold

Hypothesis Tester:
Let's form and test some hypotheses based on our current information:
1. Hypothesis: Peter is in house 2, loves classical music, and is super tall.
   Test: This fits with Alice being directly left of the classical music lover, and Peter being left of Bella's house.
2. Hypothesis: Carol is in house 3 and her mother's name is Aniya.
   Test: This fits with Carol being tall, to the right of Holly's child, and Bella (Aniya's child) being to Carol's right.
3. Hypothesis: Bob is in house 4, loves pop music, is short, and owns a cat.
   Test: This fits with Bob being Timothy's parent, loving pop music, being the cat lover, and being short.
4. Hypothesis: The person in house 1 (Alice) is very short and her mother's name is Penny.
   Test: This fits with the very short person being left of Aniya's child and two houses from the short person.

};
\end{tikzpicture}

\clearpage

\begin{tikzpicture}[remember picture, overlay]
  \draw[thick, rounded corners=15pt]
    ([shift={(1in,-1in)}]current page.north west) rectangle ([shift={(-1in,1in)}]current page.south east);
    \node[anchor=north west, text width=\paperwidth-2in, align=center] at ([shift={(1in,-1in)}]current page.north west)
    {\Large Grid Prompt};
    \node[anchor=south west, text width=\paperwidth-3in, align=left] at ([shift={(1.5in,1in)}]current page.south west)
    {\footnotesize
 Round 2:

Logical Deduction Expert:
Based on our hypotheses and constraints, we can make some concrete deductions:
1. Alice is in house 1: hip hop, very short, mother is Penny, child is Fred.
2. Peter is in house 2: classical music, super tall, mother is Holly, child is Alice.
3. Carol is in house 3: tall, country music (process of elimination), mother is Aniya, child is Bella.
4. Bob is in house 4: pop music, short, mother is Kailyn, child is Timothy, owns a cat.
5. Eric is in house 5: rock music, very tall, mother is Janelle, child is Meredith, owns a fish.
6. Arnold is in house 6: jazz music, average height, mother is Sarah, child is Samantha, owns a bird.

Data Organizer:
Let's update our table with the new information:

House | Name   | Music Genre | Mother  | Children | Height     | Animal
1     | Alice  | hip hop     | Penny   | Fred     | very short | dog
2     | Peter  | classical   | Holly   | Alice    | super tall | rabbit
3     | Carol  | country     | Aniya   | Bella    | tall       | horse
4     | Bob    | pop         | Kailyn  | Timothy  | short      | cat
5     | Eric   | rock        | Janelle | Meredith | very tall  | fish
6     | Arnold | jazz        | Sarah   | Samantha | average    | bird

Pattern Recognition Specialist:
Let's verify if all the patterns from the clues are satisfied:
1. The rabbit owner (Peter) is directly left of Aniya's child (Carol) - Satisfied
2. Holly's child (Peter) is directly left of Carol - Satisfied
3. Meredith's house (5) is to the right of Aniya's house (3) - Satisfied
4. The very short person (Alice) is left of Aniya's house (3) - Satisfied
5. Samantha's house (6) is right of the very tall person (Eric, 5) - Satisfied
6. Bella's house (3) is right of Peter's house (2) - Satisfied
7. The fish enthusiast (Eric, 5) is left of the bird keeper (Arnold, 6) - Satisfied
8. The fish enthusiast (Eric, 5) is right of Alice's child (Peter, 2) - Satisfied
9. There is one house between Bella (3) and the rock music lover (Eric, 5) - Satisfied
10. There are two houses between Penny's child (Alice, 1) and the short person (Bob, 4) - Satisfied

Constraint Satisfaction Solver:
Let's verify if all constraints are satisfied:
1. Each house has a unique person, music genre, mother's name, child's name, height, and animal - Satisfied
2. The pop music lover (Bob) is the cat lover - Satisfied
3. The jazz music lover (Arnold) has a mother named Sarah - Satisfied
4. The super tall person (Peter) has a mother named Holly - Satisfied
5. Bob is Timothy's parent - Satisfied
6. The short person (Bob) is the cat lover - Satisfied
7. Eric is the fish enthusiast - Satisfied
8. The rock music lover (Eric) has a mother named Janelle - Satisfied
9. Alice is directly left of the classical music lover (Peter) - Satisfied
10. Bella's parent (Carol) has a mother named Aniya - Satisfied
11. The hip-hop lover (Alice) is in the first house - Satisfied
12. Carol is tall - Satisfied

Sequence Analyst:
All sequences we identified earlier are satisfied in our final arrangement:
1. Alice (hip hop) - Peter - Carol - Bob (pop, short, cat) - Eric (fish) - Arnold (jazz, bird) - Confirmed
2. Very short (Alice) - Super tall (Peter) - Aniya's child (Bella/Carol) - Short (Bob) - Rock music lover (Eric) - Confirmed
3. Penny's child (Alice) - Peter - Carol - Short person/cat lover (Bob) - Confirmed
4. Person with Alice as child (Peter) - Carol (tall) - Bob - Eric - Arnold - Confirmed

Hypothesis Tester:
All our hypotheses have been confirmed and integrated into the final solution. No contradictions were found.
Conclusion: The solution satisfies all clues, constraints, and logical deductions.

Round 3:

Logical Deduction Expert:
After a thorough review, I confirm that our solution is consistent with all given clues and logical deductions. No further adjustments are necessary.

Data Organizer:
The final table represents a complete and accurate solution to the puzzle. All attributes are correctly assigned and unique for each house.

Pattern Recognition Specialist:
All patterns identified in the clues are correctly represented in our final solution. No discrepancies or contradictions are found.

Constraint Satisfaction Solver:
Every constraint derived from the clues has been satisfied in our solution. The arrangement is optimal and consistent with all given information.

Sequence Analyst:
The sequences we identified are all correctly represented in the final arrangement. The relative positions of all elements are consistent with the clues.

Hypothesis Tester:
Our initial hypotheses have been validated and successfully integrated into the final solution. No alternative arrangements satisfy all the clues as completely as our current solution.
“
\}
Solution:
House | Name   | Music Genre | Mother  | Children | Height     | Animal
1     | Alice  | hip hop     | Penny   | Fred     | very short | dog
2     | Peter  | classical   | Holly   | Alice    | super tall | rabbit
3     | Carol  | country     | Aniya   | Bella    | tall       | horse
4     | Bob    | pop         | Kailyn  | Timothy  | short      | cat
5     | Eric   | rock        | Janelle | Meredith | very tall  | fish
6     | Arnold | jazz        | Sarah   | Samantha | average    | bird

\#\#\#Instructions

You are now going to solve a logic puzzle. Use logical steps provided by the debate process with each persona above. Keep in the mind that the examples are just examples of what the persona setup should look like. Ensure to follow the rules of \{n\} personas. Think and output in increasing complexity, and verify each answer to ensure its right.

\# Puzzle to Solve

\{puzzle\}

\# Instruction

Now please solve the above puzzle with the debating technique that was shown above. Present the debating technique above within the reasoning portion.
It is extremely important you follow the following template, in json (so try to follow braces starting and closing), no matter what:

\{json\_template\}
};
\end{tikzpicture}

\end{document}